\documentclass{article} 
\usepackage{iclr2027_conference,times}

\usepackage{amsmath,amsfonts,bm}

\def\eqref#1{equation~\ref{#1}}

\def\1{\bm{1}}

\DeclareMathAlphabet{\mathsfit}{\encodingdefault}{\sfdefault}{m}{sl}
\SetMathAlphabet{\mathsfit}{bold}{\encodingdefault}{\sfdefault}{bx}{n}

\usepackage{hyperref}
\usepackage{url}
\usepackage[table]{xcolor}
\usepackage{graphicx}
\usepackage{enumitem}
\usepackage{array}
\usepackage{mathrsfs}
\usepackage{amsthm}
\usepackage{amsmath}
\usepackage{amsfonts}
\usepackage{multirow}
\usepackage{booktabs}
\usepackage{color}
\usepackage{arydshln}
\usepackage{makecell}
\usepackage{soul}
\usepackage{cleveref}
\usepackage{bbding}
\usepackage{subfigure} 
\usepackage{colortbl}
\usepackage{subcaption}
\usepackage[most]{tcolorbox}
\tcbuselibrary{listings,breakable}
\usepackage{listings}
\usepackage{wrapfig}
\usepackage{capt-of}

\usepackage[most]{tcolorbox}

\newtcolorbox{promptbox}[1]{
  enhanced,
  breakable,
  title={#1},
  colback=gray!3,
  colbacktitle=gray!12,
  colframe=black!50,
  coltitle=black,
  fonttitle=\bfseries\small,
  fontupper=\small,
  boxrule=0.5pt,
  arc=1pt,
  left=8pt,
  right=8pt,
  top=6pt,
  bottom=6pt,
  before skip=8pt,
  after skip=8pt
}

\title{MVGraph: Benchmarking LVLMs on Evidence Composition across Multi-View Visual Graph}

\author{Yingjie Zhu\textsuperscript{\rm 1,2}, Han Cheng\textsuperscript{\rm 1}, Xuefeng Bai\textsuperscript{\rm 1}, Youcheng Pan\textsuperscript{\rm 2}, Yang Xiang\textsuperscript{\rm 2}, Kehai Chen\textsuperscript{\rm 1} \\
     \textsuperscript{\rm 1}Harbin Institute of Technology, Shenzhen, China\\
     \textsuperscript{\rm 2}Peng Cheng Laboratory, Shenzhen, China\\
    \texttt{\{zhuyj,2024311560\}@stu.hit.edu.cn}, 
    \texttt{\{xiangy,panych\}@pcl.ac.cn},  \\
    \texttt{\{chenkehai,baixuefeng\}@hit.edu.cn} \\
    }

\iclrfinalcopy 
\begin{document}

\maketitle

\begin{abstract}
We introduce {\textsc{MVGraph}}, a benchmark for evaluating large vision-language models (LVLMs) on multi-graph reasoning where the evidence required by a query is distributed across multiple interrelated visual graph views. 
Although such evidence commonly resides in source-specific graphs, temporal snapshots, or local observations, existing graph reasoning benchmarks remain confined to a single self-contained structure, assessing only graph perception and reasoning while leaving the alignment and composition of evidence across views underexplored.
\textsc{MVGraph} comprises 8k instances spanning 8 \textit{graph-theoretic} tasks, abstracted from two multi-graph scenarios (partial and temporal graph views). Each problem is programmatically constructed and verified to be unsolvable from a single view.
Beyond end-to-end evaluation, we decompose multi-graph reasoning into four complementary capabilities---\textit{Perception}, \textit{Alignment}, \textit{Composition}, and \textit{Reasoning}---and design 29 diagnostic tasks with 64k instances to probe each one, along with golden intermediate representations for evaluating the intermediate process.
Across 15 LVLMs, we reveal the difficulty of multi-graph reasoning and identify cross-view edge alignment and relation-level composition as prominent bottlenecks through stage-oriented diagnostics. Meanwhile, temporal constraints and increasing view counts expose further capability boundaries.
Code and data are available at \url{https://github.com/AAAndy-Zhu/MVGraph}.

\end{abstract}

\section{Introduction}

Graphs provide a natural abstraction of relational structure in molecular property modeling \citep{pmlr-v70-gilmer17a}, web-scale recommendation \citep{ying2018graph}, traffic forecasting \citep{li2018diffusion}, and knowledge-base reasoning \citep{11647748}. Although these applications are typically formalized as computation over a single graph, the information required by a query is often distributed across multiple graphs rather than contained in one of them. It may reside in source- or relation-specific graphs that capture complementary aspects of the same system, in temporal snapshots that record how topology and attributes evolve \citep{kazemi2020representation}, or in local observations that each cover only a subset of the relevant nodes and edges. Determining the answer therefore requires establishing cross-graph correspondences and combining complementary information across graphs, rather than reasoning over a single complete graph.

Large language models (LLMs) and large vision-language models (LVLMs) have recently shown strong performance on graph-structured tasks. LLMs have achieved strong results on text-serialized graphs, including connectivity, path finding, and combinatorial optimization \citep{wang2023nlgraph,chen2024graphwiz}. 
LVLMs have further advanced this line of work by rendering graphs as visual graphs, which provide a unified interface that naturally captures relational structure while remaining compatible with multimodal graphs, visually grounded knowledge bases, and other graph types that are difficult to serialize as text \citep{das2024modality,wei2024gita,deng2024graphvis}.
Existing evaluations, however, remain largely confined to single-graph settings, where a model need only perceive one graph and reason over it. 
In contrast, a multi-graph query requires the model to parse the structure of each view, match entities or states that recur across graphs, integrate complementary evidence, and only then infer the answer. 
This leaves a critical research gap: 
\textit{Existing evaluations do not establish whether current LVLMs can recover and compose structural evidence distributed across multiple interrelated graph views when no individual view suffices.}

\begin{figure}
    \centering
    \includegraphics[width=\linewidth]{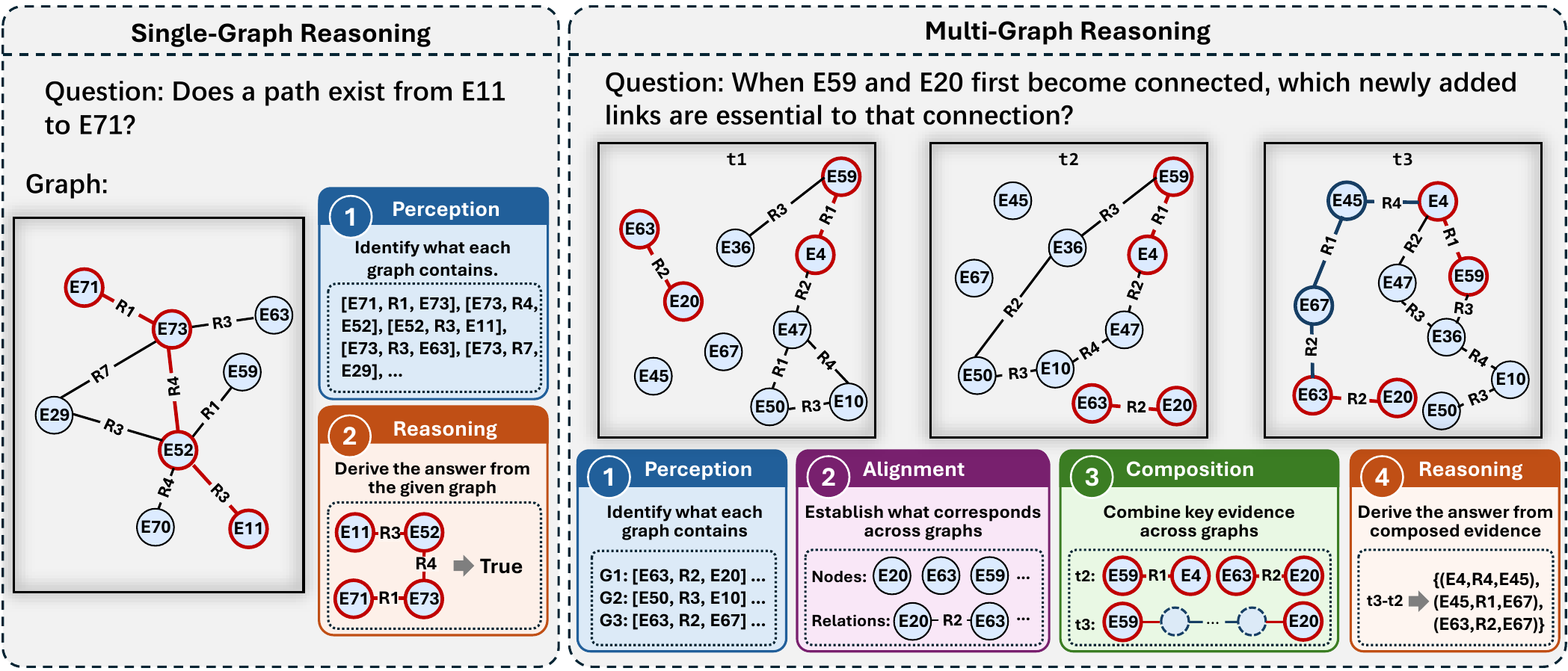}
    \caption{Illustration of evidence-fragmented visual graph reasoning in \textsc{MVGraph}, where three views provide complementary path segments.}
    \vspace{-0.8cm}
    \label{fig:introduction}
\end{figure}

To address this gap, we introduce \textbf{\textsc{MVGraph}}, a benchmark for evaluating LVLMs' graph reasoning across multiple interrelated visual graph views.
As illustrated in Figure~\ref{fig:introduction}, unlike prior benchmarks that primarily evaluate single-graph reasoning---perceiving a self-contained structure and deriving an answer from it, \textsc{MVGraph} targets multi-view reasoning, where the answer further depends on aligning recurring entities and combining complementary evidence across views.
Specifically, \textsc{MVGraph} contains 72,000 instances spanning two scenarios and 37 graph-theoretic and diagnostic tasks, with controlled variation in graph number and size, alignment ambiguity, cross-graph dependency depth, graph heterogeneity, and irrelevant evidence. 
Each instance is programmatically constructed and validated such that no individual view contains sufficient evidence to determine the answer, making task success dependent on evidence distributed across views.

To make the constituent capabilities of multi-view graph reasoning directly measurable rather than inferred from final accuracy, \textsc{MVGraph} introduces a capability-oriented evaluation schema. 
It decomposes multi-graph reasoning into four complementary capabilities: \textbf{Perception}, \textbf{Alignment}, \textbf{Composition}, and \textbf{Reasoning}. 
\textit{Perception} captures the structure and content of each graph; 
\textit{Alignment} identifies corresponding entities, relations, or states across graphs; 
\textit{Composition} organizes complementary cross-graph evidence into a coherent representation; and \textit{Reasoning} derives the answer from the integrated evidence. 
For each capability, we design targeted evaluation tasks together with a progressive evaluation that supplies verified outputs from preceding capabilities, enabling direct capability assessment and measurement of the resulting performance recovery. 
\textsc{MVGraph} thus evaluates not only overall multi-view graph reasoning, but also each constituent capability and the capability that most constrains model performance.


Across 15 LVLMs, our end-to-end evaluation shows that graph-theoretic reasoning over multiple visual views remains challenging. Through further stage-oriented diagnostics on seven LVLMs, we find that \textbf{cross-view edge alignment and relation-level composition remain significant weaknesses, despite stronger performance in identifying recurring nodes and relevant views}. Providing explicit structural evidence substantially improves performance on several partial-view graph-theoretic tasks, suggesting that \textbf{difficulties in processing distributed visual evidence} contribute to poor end-to-end performance. Tasks based on temporal sequences nevertheless remain challenging even with prepared structures, revealing an additional limitation in reasoning across snapshots.
Finally, composition performance also deteriorates markedly for several LVLMs as the number of views increases.
These findings establish MVGRAPH not only as a new evaluation resource, but also as a means to uncover capability differences obscured by existing single-graph evaluations.

Our contributions are as follows:
\begin{itemize}[itemsep=2pt,topsep=0pt,parsep=0pt,leftmargin=11pt]
\item \textbf{A comprehensive multi-graph benchmark.} We construct \textsc{MVGraph}, spanning two scenarios and 37 tasks with controlled complexity, genuine cross-graph dependencies, and gold intermediate representations.
\item \textbf{A capability-oriented diagnostic schema.} We organize multi-graph reasoning into Perception, Alignment, Composition, and Reasoning, and combine capability-level metrics with progressive evaluation to identify model bottlenecks.
\item \textbf{Empirical findings and validated measurements.}
Across 15 LVLMs, we reveal the difficulty of multi-graph reasoning and identify cross-view edge alignment and relation-level composition as prominent bottlenecks through stage-oriented diagnostics. We further characterize limitations under temporal constraints and increasing view counts, while programmatic checks and view ablations provide evidence for answer correctness and multi-view dependence.
\end{itemize}

\section{Related Work}

\paragraph{Visual Graph Benchmarking for LVLMs.}
Recent studies have begun to evaluate LVLMs on visual graph understanding and reasoning. VisionGraph \citep{li2024visiongraph} evaluates visual graph-theoretic problems, while \textsc{VGCure} \citep{zhu2025vgcure} broadens the evaluation to fundamental graph perception and reasoning. VisGraphVar \citep{sartori2024visgraphvar} studies robustness to visual variations, and DiGraphHal-Bench \citep{fan2026digraphhal} focuses on hallucination over complex directed graphs. Some benchmarks also place multiple graphs within a single image, where IsoBench \citep{fu2024isobench} includes graph isomorphism within a broader evaluation of visual and textual representations, Visual Graph Arena \citep{babaiee2025visual} tests isomorphism across different layouts, and VisGraphVar compares drawings of labeled graphs under visual style changes.
More recently, GraphVerse \citep{sun2026graphverse} extends visual graph evaluation to paired images through graph edit distance and maximum common subgraph tasks.
However, these studies do not systematically evaluate graph-theoretic problem solving when the necessary structure must be assembled from multiple complementary partial views or interpreted across ordered temporal views, together with diagnostics of the capabilities involved in that process.

\paragraph{Multi-Image Visual Reasoning.}
Recent benchmarks have extended LVLM evaluation from single-image understanding to reasoning over multiple images.
\textsc{MIRB} \citep{zhao2024mirb} and \textsc{MuirBench} \citep{wang2024muirbench} evaluate a broad range of multi-image relations, including comparison, co-reference, spatial understanding, temporal reasoning, and multi-hop inference, while \textsc{MMIU} \citep{meng2024mmiu} and \textsc{MIBench} \citep{liu2024mibench} further expand task diversity and evaluation scale across heterogeneous multi-image scenarios.
\textsc{MileBench} \citep{song2024milebench} studies long-context multimodal understanding as the number of input images increases, and reasoning-oriented benchmarks such as \textsc{ReMI} \citep{kazemi2024remi} and \textsc{MMRB} \citep{cheng2025mmrb} emphasize multi-step inference over visual evidence.
More recently, \textsc{VCBench} \citep{wang2025vcbench} and \textsc{OMIBench} \citep{chen2026omibench} construct problems in which complementary evidence is distributed across multiple images, reducing the possibility of answering from a single image alone.
These benchmarks motivate evaluating whether models genuinely use multiple images, but fail to target the graph-specific demands of recovering local topology, tracking recurring or changing graph elements, composing task-relevant structures, and solving graph-theoretic problems.
\textsc{MVGraph} studies these demands through end-to-end graph-theoretic challenges and stage-oriented tasks probing perception, alignment, composition, and reasoning when no individual view suffices.

\section{\textsc{MVGraph}: Benchmark Design and Construction}
\label{sec:benchmark}

\begin{figure}[t]
    \centering
    \includegraphics[width=\linewidth]{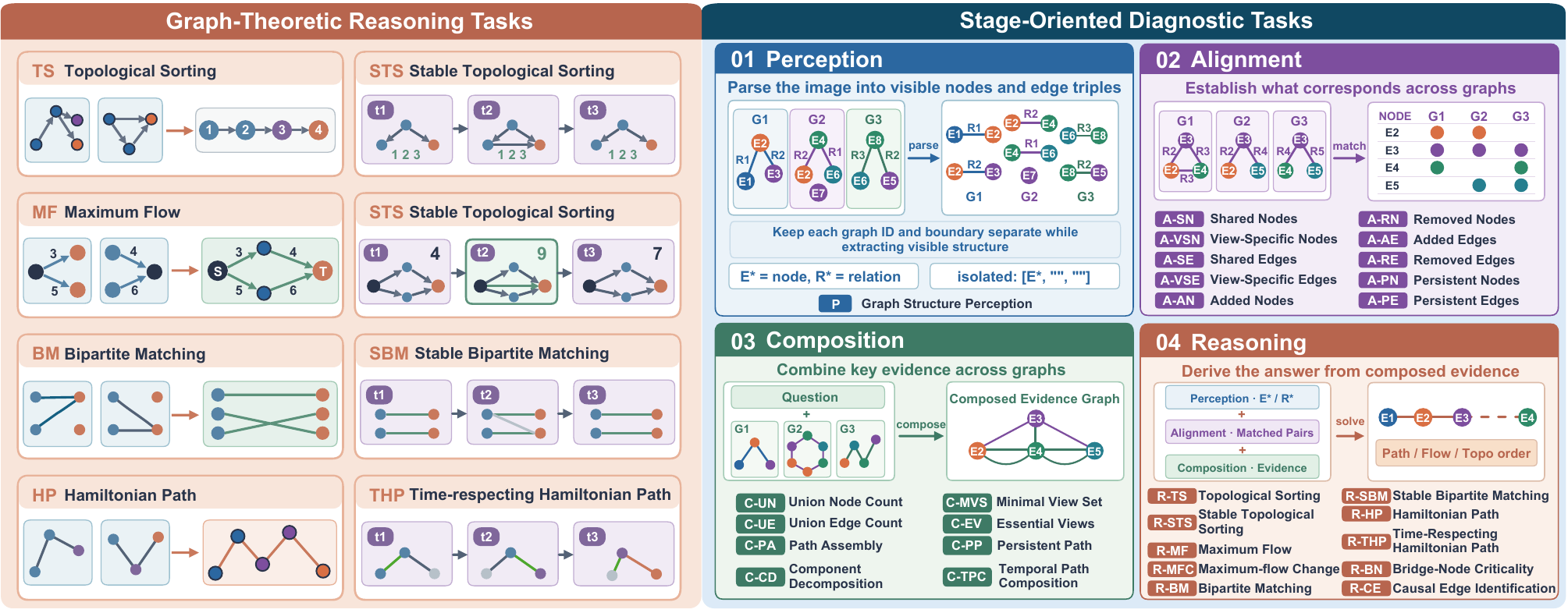}
    \caption{Overview of \textsc{MVGraph}. \textit{Left:} End-to-end graph-theoretic tasks span partial and temporal graph views. \textit{Right:} Stage-oriented diagnostic tasks probe perception, alignment, composition, and reasoning.}
    \label{fig:benchmark_overview}
\end{figure}

\textsc{MVGraph} evaluates how LVLMs reason over multiple interrelated visual graph views.
As shown in Figure~\ref{fig:benchmark_overview}, it combines end-to-end \textit{graph-theoretic} challenges with stage-oriented diagnostic tasks that probe the \textit{perception}, \textit{alignment}, \textit{composition}, and \textit{reasoning} capabilities required to solve them.
Both components are instantiated in two settings.
In \emph{partial graph views}, images expose different portions of an underlying graph, and task-relevant structure must be assembled across views.
In \emph{temporal graph views}, images depict successive graph states, and structural evidence must be interpreted in temporal order rather than merged indiscriminately.

\subsection{Task Design}
\paragraph{End-to-End Graph-Theoretic Challenges.}
The primary challenge suite consists of 8 graph-theoretic tasks across four problem families, including \textit{topological sorting}, \textit{maximum flow}, \textit{bipartite matching}, and \textit{Hamiltonian path}.
In partial graph views, models must assemble the relevant structure across images before solving the corresponding problem.
In temporal graph views, they must account for structural evolution: producing topological orders while limiting the reordering of persistent nodes, comparing maximum-flow values between successive states, finding maximum matchings while limiting changes to persistent matches, or constructing a time-respecting Hamiltonian path.
These tasks are evaluated end-to-end from graph images and cannot, particularly in the temporal setting, be reduced to reasoning over an indiscriminate union of observed edges.

\paragraph{Stage-Oriented Diagnostic Tasks.}
To examine capabilities underlying end-to-end performance, we organize diagnostic evaluation around perception, alignment, composition, and reasoning.
(I) In \emph{Perception}, models receive multiple graph images and produce a separate textual representation of the nodes and edge triples in each view.
(II) \emph{Alignment} tasks probe the identification of shared, view-specific, added, removed, and persistent graph elements across views.
(III) \emph{Composition}-oriented tasks examine whether models can organize task-relevant evidence, including assembled paths, component subgraphs, necessary view sets, and temporal path structures.
(IV) \emph{Reasoning} is assessed through structure-prepared versions of the eight graph-theoretic challenges, together with bridge-node criticality and causal edge identification.
For the perception, alignment, and composition diagnostics, models receive only the graph images and the corresponding query.
For the reasoning diagnostic, we provide task-relevant graph structure in textual form, allowing us to focus on structure reasoning without requiring visual evidence extraction.

\subsection{Instance Construction}
\paragraph{Graph Sources and View Generation.}
We construct instances from procedurally generated graphs and graph topologies derived from SmallPedia.
Procedural generation provides control over task-specific structural properties, while SmallPedia-derived topologies increase variation in the underlying graph structure.
For each instance, we select an underlying graph or sequence of graph states that satisfies the task constraints and generate multiple interrelated views from it.
Partial-view instances expose overlapping portions of an underlying graph, whereas temporal instances depict ordered states with changes in nodes or edges.
Entity identifiers remain consistent across views, but each image is laid out independently so that corresponding entities cannot be identified by position alone.
We use semantically neutral \texttt{E*} identifiers for nodes and \texttt{R*} identifiers for edge relations, where maximum-flow tasks additionally display numerical edge capacities.

\paragraph{Question and Answer Generation.}
For each task, we select graph instances satisfying its structural constraints and instantiate a question template with the relevant entities, view indices, and time steps.
Reference answers are computed from the underlying graph records using task-specific algorithms.
Diagnostic answers include graph-element sets, counts, edge-triple paths, component subgraphs, and view indices, while temporal path composition additionally requires the two component subgraphs before a connectivity transition and a path after it.
The graph-theoretic tasks require outputs such as topological orders, maximum-flow values, maximum matchings, and Hamiltonian paths, with time-indexed solutions where appropriate.
For tasks that may admit multiple valid solutions, we use task-specific programmatic verifiers to check whether each predicted solution is feasible or optimal, thereby accepting any valid solution rather than requiring an exact match to a single reference answer.

\subsection{Validation and Dataset Statistics}
\paragraph{Validation.}
We validate each instance against the structural prerequisites of its task, including acyclicity, valid edge capacities, or bipartiteness.
We then check that the reference answer computed from the underlying graph structures is consistent with the question and the selected views.
For tasks with multiple valid solutions, we verify that the stored answer satisfies the task constraints and assess model outputs by the same constraints rather than by exact agreement with one reference answer.
We additionally apply task-specific evidence-necessity checks to exclude instances that can be answered from a single view or without a required view.
Finally, we check that the rendered images preserve the intended graph elements and that their annotations are legible.

\begin{table*}[t]
\centering
\caption{Statistics of \textsc{MVGraph}.}
\label{tab:dataset_statistics}
\small
\setlength{\tabcolsep}{7pt}
\begin{tabular}{lrrrrrr}
\toprule
\textbf{Task group} &
\textbf{\# Tasks} &
\textbf{\# Instances} &
\textbf{Views} &
\textbf{Avg. views} &
\textbf{Avg. nodes} &
\textbf{Avg. edges} \\
\midrule
Graph-Theoretic & 8  & 8,000  & 2--4 & 3.06 & 14.77 & 34.03 \\
\midrule
Perception                & 1  & 36,000 & 2--8 & 4.46 & 21.47 & 26.93 \\
Alignment                 & 10 & 10,000 & 2--8 & 4.89 & 24.07 & 28.49 \\
Composition               & 8  & 8,000  & 2--8 & 4.81 & 25.33 & 24.67 \\
Reasoning                 & 10 & 10,000 & 2--8 & 3.43 & 16.19 & 30.98 \\
\midrule
\textbf{Total}             & \textbf{37} & \textbf{72,000} & \textbf{2--8}
                           & -- & -- & -- \\
\bottomrule
\end{tabular}
\end{table*}

\paragraph{Dataset Statistics.}
\textsc{MVGraph} comprises 37 evaluation tasks, including 8 end-to-end graph-theoretic challenges and 29 stage-oriented diagnostic tasks covering perception, alignment, composition, and reasoning.
Across these evaluation protocols, the benchmark contains 72,000 task-instance records, each involving 2--8 interrelated visual graph views.
The end-to-end challenges use 2--4 views, while the diagnostic tasks extend the evaluation to as many as 8 views.
Table~\ref{tab:dataset_statistics} summarizes the statistics of \textsc{MVGraph}. More details are available in Appendix \ref{app:data_construction}.

\section{Experimental Setup}
\label{sec:experimental_setup}

\paragraph{Models.}
We evaluate 15 distinct LVLMs on the eight end-to-end graph-theoretic tasks with thinking or no-thinking mode.
The stage-oriented diagnostics are conducted only with the seven thinking models, comprising four locally deployed models, including Qwen3.5-9B \citep{qwen35blog}, Gemma-4-12B-it \citep{team2026gemma}, GLM-4.6V-Flash \citep{hong2025glm}, and Ovis2.5-9B \citep{lu2025ovis2} and three API-accessed models, including Qwen3.8-Flash \citep{qiu2026design}, GLM-5.3-Flash \citep{zeng2026glm}, and DeepSeek-Flash \citep{xu2026deepseek}.
All local inference is performed using vLLM.

\paragraph{Evaluation Protocol.}
For the end-to-end graph-theoretic tasks, models receive multiple graph images, followed by the task question and output-format instruction.
The perception, alignment, and composition diagnostics likewise use visual graph inputs. In perception, multiple images are presented together and the model reconstructs the nodes and edge triples of each view separately.
For the reasoning diagnostics, models instead receive task-prepared textual graph structures and composition evidence, allowing us to assess reasoning when the relevant structure has been made explicit.
All evaluations are zero-shot.

\paragraph{Metrics.}
For perception, we report exact-match accuracy and node and edge F1 by comparing the reconstructed structure of each view with its ground-truth graph.
For all other tasks, we report exact-match based accuracy, assigning no partial credit.
Set-valued answers are compared after order-insensitive normalization, whereas paths and temporal outputs must preserve their required order.
For graph-theoretic tasks with multiple valid solutions, task-specific programmatic verifiers determine whether a prediction is valid and, where required, optimal.

\section{Results and Analysis}
We organize our study around four research questions: (RQ1) how well current LVLMs reason across multiple interrelated graphs; (RQ2) which stage limits performance, localized through progressive oracle interventions; (RQ3) how performance scales with graph views; and (RQ4) whether the benchmark reliably measures the intended capabilities.

\subsection{End-to-End Graph-Theoretic Performance}

\paragraph{Main results.}
Table~\ref{tab:graph_theoretic_thinking} reports end-to-end performance on the eight graph-theoretic tasks.
Overall, \textbf{performance remains far from saturated}.
GLM-5.3-Flash achieves the highest reported average, closely followed by Qwen3.8-Flash, while Gemma-4-12B-it leads the locally deployed models.
DeepSeek-Flash trails the other two API-accessed models, suggesting that substantially greater model scale alone does not guarantee reliable multi-graph reasoning.
\textbf{Performance also differs across the two multi-graph settings}.
The two leading LVLMs generally perform better on partial graph views than on temporal graph views.
In particular, their strong results on Bipartite Matching (BM) do not carry over to Stable Bipartite Matching (SBM) across snapshots, and Stable Topological Sorting (STS) remains difficult.
Maximum-flow Change (MFC) is an exception to this pattern, although its comparatively high scores warrant caution because the task asks models to identify a transition rather than report a numerical flow value.
Taken together, these results demonstrate substantial variation across models and graph settings, while leaving the sources of their errors unresolved.
More results and analysis are provided in the Table~\ref{tab:graph_theoretic_all} and Appendix~\ref{app:main_results}.

\begin{table*}[t]
\centering
\small
\caption{Accuracy (\%) on end-to-end graph-theoretic tasks of
\textsc{MVGraph} in thinking mode. $\dagger$ denotes models accessed via API.
The best result in each column within each model group is bolded.}
\label{tab:graph_theoretic_thinking}
\begin{tabular}{lrrrrrrrrr}
\toprule
\multirow{2}{*}{\textbf{Models}}
& \multicolumn{4}{c}{\textbf{Partial graph views}}
& \multicolumn{4}{c}{\textbf{Temporal graph views}}
& \multirow{2}{*}{\textbf{Avg.}} \\
\cmidrule(lr){2-5}\cmidrule(lr){6-9}
& \textbf{TS} & \textbf{MF} & \textbf{BM} & \textbf{HP}
& \textbf{STS} & \textbf{MFC} & \textbf{SBM} & \textbf{THP} & \\
\midrule
Qwen3.5-9B
& \textbf{0.1} & 0.3 & 36.9 & 40.0
& 0.0 & 30.2 & 0.2 & 0.4 & 13.5 \\
Gemma-4-12B-it
& 0.0 & \textbf{5.5} & \textbf{60.6} & \textbf{43.1}
& \textbf{0.1} & 45.3 & \textbf{0.4} & \textbf{8.7}
& \textbf{20.5} \\
GLM-4.6V-Flash
& 0.0 & 0.7 & 7.2 & 4.7
& 0.0 & \textbf{46.5} & 0.0 & 0.0 & 7.4 \\
Ovis2.5-9B
& 0.0 & 0.2 & 18.5 & 2.3
& 0.0 & 41.3 & 0.0 & 0.2 & 7.8 \\
\midrule
Qwen3.8-Flash$^\dagger$
& \textbf{35.0} & \textbf{54.0} & \textbf{97.0} & 77.0
& 8.0 & \textbf{59.0} & 8.0 & 65.0 & 50.4 \\
GLM-5.3-Flash$^\dagger$
& 33.0 & 41.0 & 95.0 & \textbf{84.0}
& \textbf{12.0} & \textbf{59.0} & \textbf{16.0} & \textbf{79.0}
& \textbf{52.4} \\
DeepSeek-Flash$^\dagger$
& 0.0 & 15.0 & 62.0 & 50.0
& 0.0 & 5.0 & 0.0 & 7.0 & 17.4 \\
\bottomrule
\end{tabular}
\end{table*}

\begin{figure*}[t]
\centering
\begin{minipage}[t]{0.485\textwidth}
    \centering
    \includegraphics[width=\linewidth]{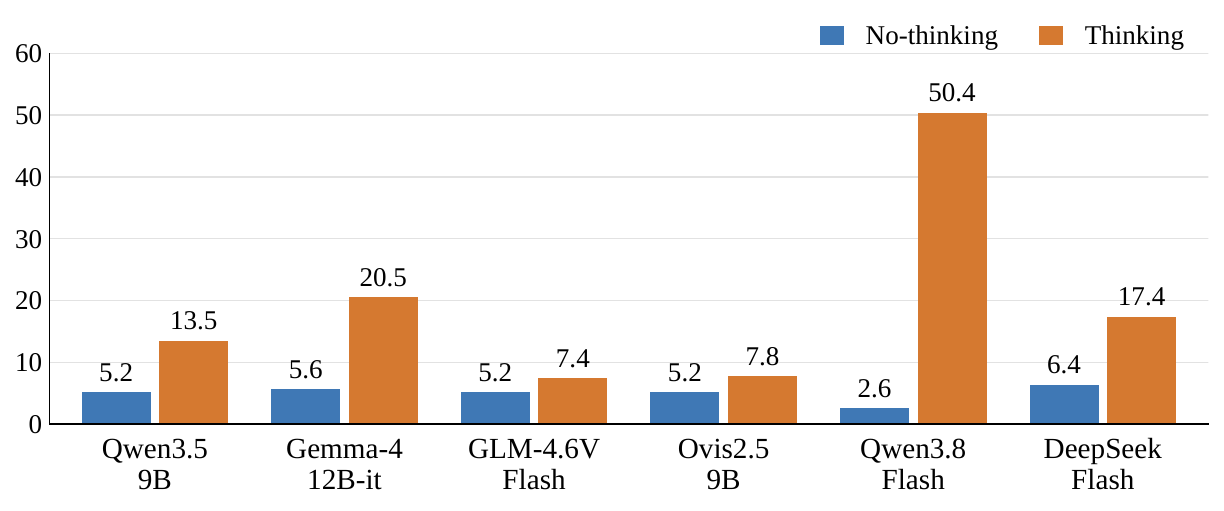}
    \caption{Average accuracy on graph-theoretic tasks with and without thinking.}
    \label{fig:thinking_gain}
\end{minipage}
\hfill
\begin{minipage}[t]{0.485\textwidth}
    \centering
    \includegraphics[width=\linewidth]{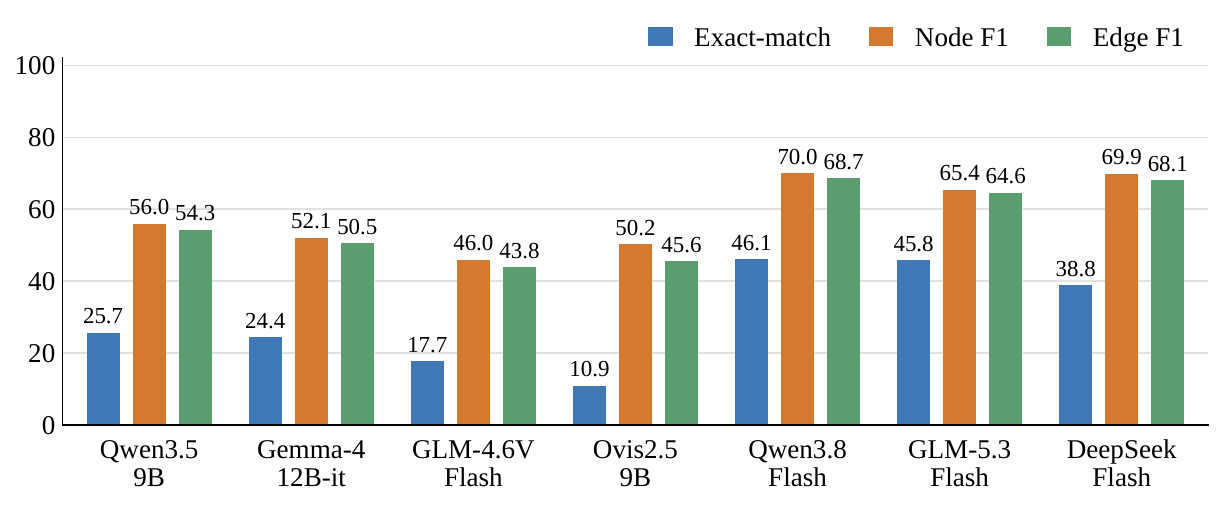}
    \caption{Perception performance of LVLMs in thinking mode.}
    \label{fig:perception_metrics}
\end{minipage}
\end{figure*}

\paragraph{Effect of thinking.}
Figure~\ref{fig:thinking_gain} compares the six models in both inference settings.
Without thinking, accuracies cluster between 2.6\% and 6.4\%; thinking improves every model, most dramatically Qwen3.8-Flash (2.6\% to 50.4\%), whereas GLM-4.6V-Flash and Ovis2.5-9B improve only modestly.
Test-time reasoning thus offers substantial but highly model-dependent gains, and the benchmark remains challenging.

\subsection{Diagnosing the Four Capabilities}
\label{sec:capability_diagnostics}
End-to-end results do not reveal which capabilities constrain multi-graph reasoning. We therefore evaluate seven representitive LVLMs on diagnostic tasks targeting perception, alignment, composition, and reasoning. We use thinking mode throughout to assess these capabilities under the stronger inference setting observed in our end-to-end experiments. 

\paragraph{Perception.}
Figure~\ref{fig:perception_metrics} shows that complete reconstruction of multiple graph views remains difficult even in thinking mode, with no model exceeding 47\% exact-match accuracy. Node F1 consistently exceeds Edge F1, suggesting that extracting edge triples poses an additional challenge.\textbf{ The gap between F1 and exact match indicates that LVLMs often recover part of the structure without reconstructing every view completely.} These results establish the difficulty of visual structure extraction before we examine the cross-view alignment and composition capabilities required specifically for multi-graph reasoning.

\begin{table*}[t]
\centering
\caption{Accuracy (\%) of thinking-mode LVLMs on the ten alignment tasks of \textsc{MVGraph}. $\dagger$ denotes models accessed via API. The best result in each column is bolded.}
\label{tab:alignment_thinking}
\scriptsize
\setlength{\tabcolsep}{3pt}
\begin{tabular}{l*{11}{r}}
\toprule
\multirow{2.5}{*}{\textbf{Models}}
& \multicolumn{4}{c}{\textbf{Partial}}
& \multicolumn{6}{c}{\textbf{Temporal}} & \multirow{2.5}{*}{\textbf{Avg.}}  \\
\cmidrule(lr){2-5}\cmidrule(lr){6-11}

& \textbf{A-SN} & \textbf{A-VSN} & \textbf{A-SE} & \textbf{A-VSE}
& \textbf{A-AN} & \textbf{A-RN} & \textbf{A-AE} & \textbf{A-RE}
& \textbf{A-PN} & \textbf{A-PE} &  \\
\midrule
Qwen3.5-9B
& 82.1 & 94.0 & 55.2 & 36.0 & 98.1 & 92.1 & 29.8 & 35.5 & 99.6 & 75.5 & 69.8 \\
Gemma-4-12B-it
& 81.3 & 91.0 & 57.1 & 29.8 & 98.4 & 91.9 & 46.5 & 44.9 & 95.0 & 80.6 & 71.7 \\
GLM-4.6V-Flash
& 97.9 & 68.0 & 58.8 & 11.0 & 81.7 & 59.9 & 18.4 & 19.1 & 96.9 & 62.9 & 57.5 \\
Ovis2.5-9B
& 69.4 & 65.7 & 53.4 & 4.0 & 86.1 & 72.0 & 8.3 & 12.6 & 92.2 & 65.9 & 53.0 \\
\midrule
Qwen3.8-Flash$^\dagger$
& \textbf{99.0} & \textbf{99.5} & 74.5 & \textbf{57.5} & 99.5 & \textbf{99.0} & 63.0 & 65.0 & \textbf{100.0} & \textbf{89.5} & 84.7 \\
GLM-5.3-Flash$^\dagger$
& 97.5 & 99.0 & \textbf{85.5} & 54.0 & 99.5 & 97.0 & 65.0 & \textbf{65.5} & \textbf{100.0} & 88.5 & \textbf{85.2} \\
DeepSeek-Flash$^\dagger$
& 98.5 & 97.5 & 31.5 & 28.0 & \textbf{100.0} & 97.0 & 45.5 & 56.5 & \textbf{100.0} & 79.5 & 73.4 \\
\bottomrule
\end{tabular}
\end{table*}

\paragraph{Alignment.}
Table~\ref{tab:alignment_thinking} reveals a pronounced gap between node-level  (A-SN, A-VSN, A-AN, A-RN, and A-PN) and edge-level (A-SE, A-VSE, A-AE, A-RE, and A-PE) alignment across all evaluated models. Shared, view-specific, added, removed, and persistent nodes are consistently identified more accurately than their edge counterparts. The gap is especially evident for view-specific edges, on which even the strongest model remains below 60\% accuracy. 
This difficulty persists despite consistent entity identifiers across views, suggesting that \textbf{tracking how relations correspond or change across views is substantially harder than aligning nodes. Since such edge-level alignment is essential for integrating structural evidence, its weakness may limit LVLMs' multi-graph reasoning performance.}

\begin{table*}[t]
\centering
\scriptsize

\setlength{\tabcolsep}{3.5pt}
\caption{Accuracy (\%) of thinking-mode LVLMs on the eight composition tasks in \textsc{MVGraph}. $\dagger$ denotes models accessed via API. The best result in each column is bolded.}
\label{tab:composition_thinking}
\begin{tabular}{lrrrrrrrrr}
\toprule
\multirow{2.5}{*}{\textbf{Models}}
& \multicolumn{6}{c}{\textbf{Partial}} & \multicolumn{2}{c}{\textbf{Temporal}} & \multirow{2.5}{*}{\textbf{Avg.}}\\
\cmidrule(lr){2-7}\cmidrule(lr){8-9}
& \textbf{C-UN} & \textbf{C-UE} & \textbf{C-PA} & \textbf{C-CD}
& \textbf{C-MVS} & \textbf{C-EV}
& \textbf{C-PP} & \textbf{C-TPC} & \\
\midrule
Qwen3.5-9B       & 94.1 & 33.7 & 95.0 & 77.5 & 99.1 & 99.2 & 98.0 & 58.5 & 81.9 \\
Gemma-4-12B-it   & 92.0 & 43.1 & 91.0 & 72.0 & 91.3 & 90.7 & 98.5 & 36.0 & 76.8 \\
GLM-4.6V-Flash   & 68.0 & 21.9 & 58.5 & 32.5 & 70.4 & 69.3 & 96.5 & 6.5  & 53.0 \\
Ovis2.5-9B       & 60.6 & 16.8 & 40.0 & 40.0 & 66.3 & 63.0 & 96.5 & 1.0  & 48.0 \\
\midrule
Qwen3.8-Flash$^\dagger$
                 & \textbf{99.0} & 67.0 & \textbf{98.0} & \textbf{89.5}
                 & 99.5 & 99.5 & \textbf{100.0} & \textbf{87.0} & \textbf{92.4} \\
GLM-5.3-Flash$^\dagger$
                 & 98.0 & \textbf{73.0} & \textbf{98.0} & 83.5
                 & \textbf{100.0} & \textbf{100.0} & \textbf{100.0} & 73.5 & 90.8 \\
DeepSeek-Flash$^\dagger$
                 & 95.0 & 26.5 & 95.0 & 85.0 & 92.0 & 89.5 & 99.5 & 81.5 & 83.0 \\
\bottomrule
\end{tabular}
\end{table*}

\paragraph{Composition.}
Table~\ref{tab:composition_thinking} shows that composition difficulty depends strongly on the structural evidence that must be integrated. Models perform well on Minimal View Set (C-MVS) and Essential Views (C-EV), indicating that they can often identify which views contribute to a connection. Yet this ability does not consistently extend to composing the relations within those views. Union Edge Count (C-UE) is substantially less accurate than Union Node Count (C-UN) for every model, with even the strongest models falling below 75\% on C-UE. A similar contrast appears in the temporal setting. Persistent Path (C-PP) is solved reliably, whereas Temporal Path Composition (C-TPC), which requires reconstructing components before a transition and a connecting path afterward, is markedly more difficult. 
\textbf{Therefore, selecting relevant views is easier than faithfully combining their edge-level evidence into a complete cross-view structure, which may in turn limit graph-theoretic reasoning over distributed evidence.}

\begin{table*}[t]
\centering
\setlength{\tabcolsep}{3pt}
\caption{Accuracy (\%) of thinking-mode LVLMs on the ten structure-prepared reasoning tasks in \textsc{MVGraph}. Models receive textualized structural evidence instead of graph images. $\dagger$ denotes models accessed via API. The best result in each column is bolded.}
\label{tab:reasoning_thinking}
\resizebox{\linewidth}{!}{
\begin{tabular}{lrrrrrrrrrrr}
\toprule
\multirow{2.5}{*}{\textbf{Models}}
& \multicolumn{5}{c}{\textbf{Partial}} & \multicolumn{5}{c}{\textbf{Temporal}} & \multirow{2.5}{*}{\textbf{Avg.}}\\
\cmidrule(lr){2-6}\cmidrule(lr){7-11}
& \textbf{R-TS} & \textbf{R-MF} & \textbf{R-BM} & \textbf{R-HP} & \textbf{R-BN}
& \textbf{R-STS} & \textbf{R-MFC} & \textbf{R-SBM} & \textbf{R-THP} & \textbf{R-CE}
&  \\
\midrule
Qwen3.5-9B
& 72.0 & 62.0 & 79.0 & 87.0 & 93.0
& 0.0 & 46.0 & 0.0 & 1.0 & 97.0 & 53.7 \\
Gemma-4-12B-it
& 91.0 & 41.0 & 71.0 & 89.0 & 97.0
& 0.0 & 67.0 & 5.0 & 44.0 & 96.0 & 60.1 \\
GLM-4.6V-Flash
& 0.0 & 2.0 & 9.0 & 3.0 & 50.5
& 0.0 & 52.0 & 0.0 & 0.0 & 65.5 & 18.2 \\
Ovis2.5-9B
& 2.0 & 14.0 & 25.0 & 5.0 & 56.5
& 0.0 & 34.0 & 0.0 & 0.0 & 62.5 & 19.9 \\
\midrule
Qwen3.8-Flash$^\dagger$
& 93.0 & 73.0 & 93.0 & 86.0 & 87.0
& 16.0 & 70.0 & 28.0 & 65.0 & 69.5 & 68.1 \\
GLM-5.3-Flash$^\dagger$
& 95.0 & 87.0 & 95.0 & 98.0 & 98.5
& \textbf{44.0} & 89.0 & \textbf{55.0} & \textbf{84.0} & 96.5 & \textbf{84.2} \\
DeepSeek-Flash$^\dagger$
& \textbf{100.0} & \textbf{100.0} & \textbf{100.0} & \textbf{100.0} & \textbf{100.0}
& 19.0 & \textbf{99.0} & 32.0 & 66.0 & \textbf{100.0} & 81.6 \\
\bottomrule
\end{tabular}}
\end{table*}

\paragraph{Reasoning.}
Unlike the preceding diagnostics, this condition provides textualized graph structures with prepared composition evidence. Table~\ref{tab:reasoning_thinking} shows that several LVLMs can solve the partial-view graph-theoretic challenges when the relevant structure is made explicit. DeepSeek-Flash, for example, solves all such challenges in this condition despite performing substantially worse on their image-based counterparts. This contrast indicates that \textbf{weak end-to-end performance cannot be attributed to graph-theoretic inference alone} and is consistent with the cross-view alignment and composition difficulties identified above. Furthermore, Stable Topological Sorting (R-STS) and Stable Bipartite Matching (R-SBM) remain difficult even with prepared structures, \textbf{revealing an additional limitation in reasoning over solutions that must remain consistent across time.}

\paragraph{Summary.}
The four diagnostics show that multi-graph reasoning is not limited by visual perception alone. Models often identify recurring nodes and relevant views, but \textbf{struggle to align edges and compose their relations into a coherent cross-view structure.} Temporal tasks place further demands on this capability by \textbf{requiring models to align which relations persist or change across snapshots.} Although explicit structural evidence improves performance on several graph-theoretic tasks, the remaining difficulty of temporally coupled problems shows that reasoning across time is also challenging.

\begin{figure*}[t]
\centering
\includegraphics[width=\linewidth]{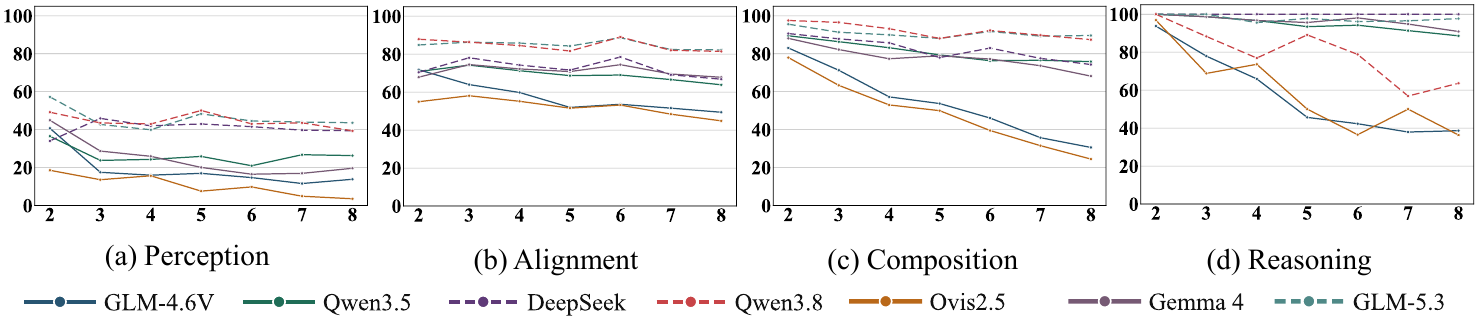}
\caption{Diagnostic performance with thinking-mode across various number of graph views. The reasoning includes only Bridge-Node Criticality (R-BN) and Causal Edge Identification (R-CE).}
\label{fig:graph_count}
\end{figure*}

\subsection{Scaling with the Number of Views.}
Figure~\ref{fig:graph_count} examines diagnostic performance as the number of graph views increases. Perception exact match generally declines when models must reconstruct more views jointly. The most pronounced degradation appears in composition, where several locally deployed models lose substantial accuracy as structural evidence is spread across more views, while the API-accessed models remain comparatively stable. Aggregate alignment performance changes less consistently with view count, although the task-level results above reveal persistent difficulty in aligning edges. These patterns suggest that faithfully integrating relation-level evidence across many views is a particularly important capacity boundary for current LVLMs.

\subsection{Benchmark Validation}

\paragraph{Programmatic verification.}
\textsc{MVGraph} is constructed using task-specific graph-generation programs and question templates, rather than manually assigned answers. Each generated instance is checked against the structural constraints of its task, including valid edge endpoints and applicable properties such as acyclicity or bipartite structure. Its reference answer is then computed from the underlying graph structures and verified by a task-specific program. For tasks with multiple valid solutions, the verifier checks whether a candidate satisfies the required structural constraints and, where applicable, achieves the optimum, rather than requiring it to match a single reference string. These checks ensure that the stored answers are correct with respect to the generated graphs and the specified task definitions.

\begin{figure}
    \centering
    \includegraphics[width=0.7\linewidth]{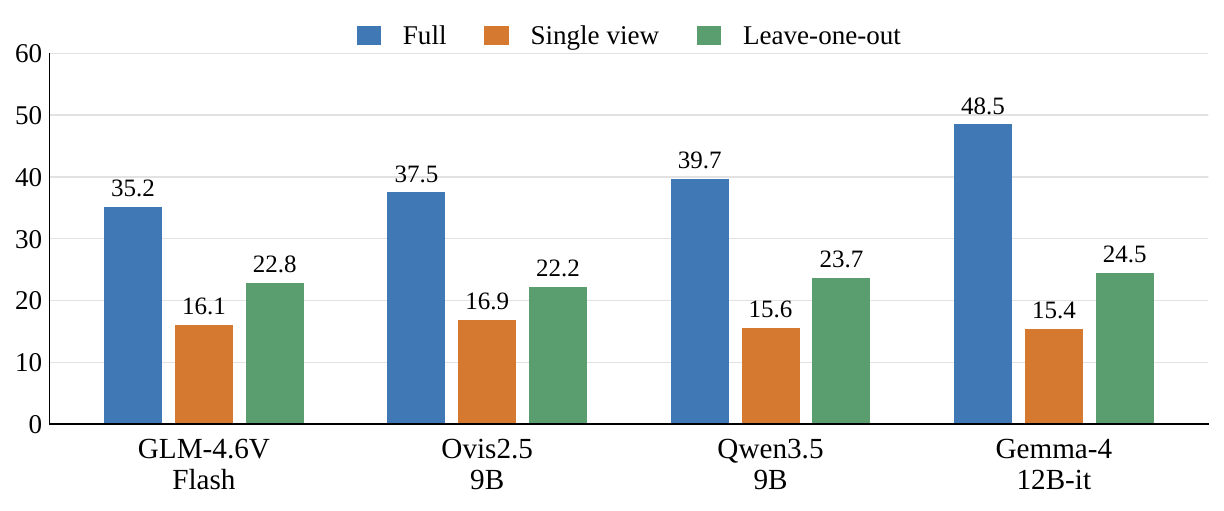}
    \caption{Effect of view availability on \textsc{MVGraph} performance, where \textit{Full} uses all graph views, \textit{Single View} retains one randomly selected view, and\textit{ Leave-One-Out} removes that same view. Bars show mean accuracy on the sampled instances.}
    \label{fig:view_ablation}
\end{figure}

\paragraph{View-ablation protocol.}
We further examine whether answering the questions benefits from access to multiple graph views. Specifically, we randomly select a subset of \textsc{MVGraph} instances and evaluate four LVLMs under three paired input conditions without thinking. \emph{Full} presents every graph view. For each instance, we randomly choose one view $I_j$; \emph{Single View} presents only $I_j$, whereas \emph{Leave-One-Out} presents every view except $I_j$. The selected index is held fixed across models, and the question and evaluation procedure remain unchanged across the three conditions.

\paragraph{Results.}
Figure~\ref{fig:view_ablation} shows a consistent ordering across all test LVLMs, with \textit{Full} outperforming \textit{Leave-One-Out}, which in turn outperforms \textit{Single View}. Accuracy drops sharply when only one graph is available and remains substantially below \textit{Full} even when just one view is removed. For the strongest model, omitting a view nearly halves its accuracy. Moreover, model performance is much more similar under the reduced-view conditions than under \textit{Full}, suggesting that their ability to exploit distributed evidence becomes most apparent when the complete set of views is available. These results provide behavioral evidence of multi-view dependence at the dataset level, complementing the programmatic checks of answer correctness.

\section{Conclusion}
We introduced \textsc{MVGraph} to evaluate reasoning over structural evidence distributed across multiple interrelated visual graph views. Its graph-theoretic challenges and stage-oriented diagnostics reveal that current LVLMs struggle not only to perceive graph structure, but also to align edge relations and compose evidence across views. Providing explicit structural information improves several partial-view tasks, yet temporally coupled reasoning remains difficult, and composition performance declines for several models as the number of views grows. These findings highlight relation-level alignment, faithful evidence composition, and temporal reasoning as important directions for improving multi-graph understanding.

\section*{AI Use Statement.}
We used ChatGPT to assist with language polishing and improving the clarity of the manuscript. The authors reviewed and take responsibility for the final content.

\bibliography{iclr2027_conference}
\bibliographystyle{iclr2027_conference}
\newpage
\appendix

\section{Details of Dataset Construction}
\label{app:data_construction}

\begin{table*}[t]
\centering
\caption{Examples of the eight end-to-end graph-theoretic reasoning tasks. For tasks admitting multiple valid solutions, one valid
answer is shown. Long structured answers are abbreviated for readability.}
\label{tab:advanced_task_examples}

\scriptsize
\setlength{\tabcolsep}{4pt}
\renewcommand{\arraystretch}{1.12}

\begin{tabular}{
    @{}
    m{0.035\textwidth}
    m{0.19\textwidth}
    m{0.42\textwidth}
    m{0.27\textwidth}
    @{}
}
\toprule
\textbf{ID} & \textbf{Task} & \textbf{Example question} &
\textbf{Example answer} \\
\midrule

\rowcolor{gray!10}
\multicolumn{4}{@{}l}{\textbf{Partial graph views}} \\
\midrule

\rowcolor{white}
TS
& Topological Sorting
& Merge all views into a directed acyclic graph and return a valid topological ordering of every node.
& \shortstack[l]{
    [E13, E9, E63, E11,\\
    E4, E10, E3, E18,\\
    E8, E48, E43, E53]
} \\
\midrule

MF
& Maximum Flow
& After merging all capacity-network views, what is the maximum flow from source \texttt{E4} to sink \texttt{E6}?
& 46 \\
\midrule

BM
& Bipartite Matching
& Given the displayed applicant and job partitions, return a maximum-cardinality matching in the merged graph.
& \shortstack[l]{
    [(E11, E2), (E33, E8),\\
    (E35, E3), (E43, E80),\\
    (E55, E38), (E58, E1)]
} \\
\midrule

HP
& Hamiltonian Path
& Merge all views and return a path that visits every node in the resulting graph exactly once.
& \shortstack[l]{
    [E9, E11, E67, E63, E13,\\
    E5, E70, E61, E12, E1,\\
    E7, E52, E80, E32]
} \\
\midrule
\midrule

\rowcolor{gray!10}
\multicolumn{4}{@{}l}{\textbf{Temporal graph views}} \\
\midrule

\rowcolor{white}
STS
& Stable Topological Sorting
& Return a valid topological order at every time step while minimizing inversions among nodes shared by consecutive snapshots.
& \shortstack[l]{
    $\pi_1$: [E19, E8, E3, E48,\\
    \phantom{$\pi_1$: [}E9, E11, E37, E1]\\
    $\pi_2$: [E54, E8, E3, E48,\\
    \phantom{$\pi_2$: [}E11, E1, E9, E37]
} \\
\midrule

MFC
& Maximum-flow Change
& Between which consecutive snapshots does the maximum flow from \texttt{E5} to \texttt{E6} increase the most?
& [2, 3] \\
\midrule

SBM
& Stable Bipartite Matching
& Return a maximum-cardinality matching at every time step while minimizing assignment changes for stable applicants.
& \shortstack[l]{
    $M_1$: [(E34, E12), \ldots,\\
    \phantom{$M_1$: [}(E26, E49)]\\
    $M_2$: [(E34, E12), \ldots,\\
    \phantom{$M_2$: [}(E68, E24)]
} \\
\midrule

THP
& Time-respecting Hamiltonian Path
& Return a Hamiltonian path and the nondecreasing image index used by each path edge, using every time step.
& \shortstack[l]{
    Path: [E14, E43, E6, E57,\\
    E8, E21, E20, E9, E36,\\
    E34, E2, E12, E51, E26]\\
    Images: [1,1,1,1,1,1,1,\\
    2,2,2,2,2,2]
} \\

\bottomrule
\end{tabular}
\end{table*}


\begin{table*}[t]
\centering
\caption{Examples of the ten Alignment tasks. For tasks admitting multiple valid solutions, one valid answer is shown. Long structured answers are abbreviated for readability.}
\label{tab:alignment_task_examples}

\scriptsize
\setlength{\tabcolsep}{4pt}
\renewcommand{\arraystretch}{1.12}

\begin{tabular}{
    @{}
    m{0.06\textwidth}
    m{0.185\textwidth}
    m{0.42\textwidth}
    m{0.27\textwidth}
    @{}
}
\toprule
\textbf{ID} & \textbf{Task} & \textbf{Example question} &
\textbf{Example answer} \\
\midrule

\rowcolor{gray!10}
\multicolumn{4}{@{}l}{\textbf{Partial graph views}} \\
\midrule

\rowcolor{white}
A-SN
& Shared Nodes
& Identify the node labels that are present in all graph views.
& [E4, E61] \\
\midrule

A-VSN
& View-Specific Nodes
& List the nodes that are exclusive to Graph 3.
& [E30] \\
\midrule

A-SE
& Shared Edges
& What edge triples are shared by every graph view?
& [[E11, R1, E56]] \\
\midrule

A-VSE
& View-Specific Edges
& What edge triples are unique to Graph 1 across the given graph views?
& \shortstack[l]{
    [[E1, R4, E61], [E4, R1, E71],\\{}
    [E15, R2, E61]]
} \\
\midrule

\rowcolor{gray!10}
\multicolumn{4}{@{}l}{\textbf{Temporal graph views}} \\
\midrule

\rowcolor{white}
A-AN
& Added Nodes
& Identify the nodes newly appearing between $t_2$ and $t_3$.
& [E1] \\
\midrule

A-RN
& Removed Nodes
& Identify the nodes disappearing between $t_1$ and $t_2$.
& [E37] \\
\midrule

A-AE
& Added Edges
& Identify the edge triples newly appearing between $t_3$ and $t_4$.
& \shortstack[l]{
    [[E4, R2, E65], [E29, R3, E38],\\{}
    [E47, R1, E65]]
} \\
\midrule

A-RE
& Removed Edges
& Which edge triples are present in $t_3$ but absent in $t_4$?
& [[E28, R1, E61], [E59, R4, E71]] \\
\midrule

A-PN
& Persistent Nodes
& List the nodes that remain present throughout the whole sequence.
& [E31, E59] \\
\midrule

A-PE
& Persistent Edges
& Which edge triples are present at every time step?
& [[E20, R1, E71]] \\

\bottomrule
\end{tabular}
\end{table*}

\begin{table*}[t]
\centering
\caption{Examples of the eight Composition tasks. For tasks admitting multiple valid solutions, one valid answer is shown. Long structured answers are abbreviated for readability.}
\label{tab:composition_task_examples}

\scriptsize
\setlength{\tabcolsep}{4pt}
\renewcommand{\arraystretch}{1.12}

\begin{tabular}{
    @{}
    m{0.06\textwidth}
    m{0.20\textwidth}
    m{0.40\textwidth}
    m{0.27\textwidth}
    @{}
}
\toprule
\textbf{ID} & \textbf{Task} & \textbf{Example question} &
\textbf{Example answer} \\
\midrule

\rowcolor{gray!10}
\multicolumn{4}{@{}l}{\textbf{Partial graph views}} \\
\midrule

\rowcolor{white}
C-UN
& Union Node Count
& How many unique node labels appear in the union of all graph views?
& 22 \\
\midrule

C-UE
& Union Edge Count
& Across all graph views, how many different edge triples are shown?
& 28 \\
\midrule

C-PA
& Path Assembly
& After merging all graph views, give one path from E28 to E21 as an ordered list of edge triples.
& \shortstack[l]{
    [[E28, R1, E77], [E77, R2, E58],\\ {} [E58, R3, E13],
    [E13, R4, E50],\\{} [E50, R1, E78], [E78, R2, E21]]
} \\
\midrule

C-CD
& Component Decomposition
& After merging all graph views, list every connected component with its nodes and edge triples.
& \shortstack[l]{
    component: \\
    nodes [E13,E19,E22,E27,\\
    E28,E38,E44];\\
    edges [[E13,R1,E27], ...]\\
    component: \\
    nodes [E26,E45,E58,E78];\\
    edges [[E26,R3,E45], ...]
} \\
\midrule

C-MVS
& Minimal View Set
& Which graph views form the smallest sufficient set for connecting E5 to E76?
& [1, 2] \\
\midrule

C-EV
& Essential Views
& Which graph views are indispensable for making E74 reachable from E55?
& [2, 3, 4] \\

\midrule
\rowcolor{gray!10}
\multicolumn{4}{@{}l}{\textbf{Temporal graph views}} \\
\midrule

\rowcolor{white}
C-PP
& Persistent Path
& In the node-and-edge intersection graph, give one path from E23 to E70.
& \shortstack[l]{
    [[E23, R1, E79], [E79, R2, E16],\\{}
    [E16, R3, E71], [E71, R4, E70]]
} \\
\midrule

C-TPC
& Temporal Path Composition
& Find adjacent times where E60 and E53 change from disconnected to connected. Report the two prior components and the later path.
& \shortstack[l]{
    $t_1$:  \\ subgraph1 [[E27,R2,E48], ...];\\
    subgraph2 [[E20,R4,E76], ...];\\
    $t_2$:  \\
    path [[E60,R1,E75],..., [E76,R1,E53]]
} \\

\bottomrule
\end{tabular}
\end{table*}

\begin{table*}[t]
\centering
\caption{Examples of the ten reasoning tasks. For tasks admitting multiple valid solutions, one valid answer is shown. Long structured answers are abbreviated for readability.}
\label{tab:pure_reasoning_task_examples}

\scriptsize
\setlength{\tabcolsep}{4pt}
\renewcommand{\arraystretch}{1.12}

\begin{tabular}{
    @{}
    m{0.06\textwidth}
    m{0.19\textwidth}
    m{0.42\textwidth}
    m{0.27\textwidth}
    @{}
}
\toprule
\textbf{ID} & \textbf{Task} & \textbf{Example question} &
\textbf{Example answer} \\
\midrule

\rowcolor{gray!10}
\multicolumn{4}{@{}l}{\textbf{Partial graph views}} \\
\midrule

\rowcolor{white}
R-TS
& Topological Sorting
& Merge all views into a directed acyclic graph and return a valid topological ordering of every node.
& \shortstack[l]{
    [E45, E63, E72, E22, E31, E65, E47,\\
    E48, E7, E30, E44, E49, E24, E62]
} \\
\midrule

R-MF
& Maximum Flow
& After merging all capacity-network views, what is the maximum flow from source \texttt{E37} to sink \texttt{E79}?
& 55 \\
\midrule

R-BM
& Bipartite Matching
& Given the applicant and job partitions in the graph facts, return a maximum-cardinality matching in the merged graph.
& \shortstack[l]{
    [(E24, E11), (E39, E3), (E5, E6), \\ 
    (E51, E17), (E62, E35), (E63, E47), \\
    (E66, E28), (E76, E38)]
} \\
\midrule

R-BN
& Bridge-Node Criticality
& Which cross-view node(s) act as necessary bridges on the route from \texttt{E60} to \texttt{E47}?
& [E28, E31, E75] \\
\midrule

R-HP
& Hamiltonian Path
& Merge the graph facts into one undirected graph and return a path that visits every node exactly once.
& \shortstack[l]{
    [E10, E16, E59, E33, E5, E22, E72, \\
    E3, E27, E49, E28, E39, E50, E73, \\ 
    E18, E55, E30]
} \\
\midrule
\midrule
\rowcolor{gray!10}
\multicolumn{4}{@{}l}{\textbf{Temporal graph views}} \\
\midrule

\rowcolor{white}
R-STS
& Stable Topological Sorting
& Give one valid topological ordering for each time step, in image order. Among all valid sequences, minimize the total number of pairwise order inversions for nodes shared by consecutive time steps.
& \shortstack[l]{
    $\pi_1$: [E79, E24, ..., E5, E40, E45]\\
    $\pi_2$: [E66, E23, ..., E25, E40, E45]\\
    $\pi_3$: [E79, E40, ..., E24, E45, E59]\\
    $\pi_4$: [E40, E21, ..., E45, E79, E23]
} \\
\midrule

R-MFC
& Maximum-flow Change
& The source is \texttt{E74} and the sink is \texttt{E54}. Between which consecutive graph images does the maximum-flow value increase the most? Return the two image numbers in chronological order.
& [2, 3] \\
\midrule

R-SBM
& Stable Bipartite Matching
& Return a maximum-cardinality matching for every time step. Among all such sequences, minimize how many stable applicants change their assigned job between consecutive images.
& \shortstack[l]{
    $M_1$: [(E39, E46), (E44, E54), ...]\\
    $M_2$: [(E39, E54), (E44, E62), ...]\\
    $M_3$: [(E39, E1), (E44, E18), ...]\\
    $M_4$: [(E39, E2), (E44, E18), ...]
} \\
\midrule

R-THP
& Time-respecting Hamiltonian Path
& Find a time-respecting Hamiltonian path that visits every node exactly once. For each path edge, give the graph image in which it is used; image numbers must be nondecreasing and every time step must be used.
& \shortstack[l]{
    Path: [E47, E73, ..., E48, E65]\\
    Images: [1,1,1,1,2,2,...,3,4,4,4,4]
} \\
\midrule

R-CE
& Causal Edge Identification
& Given the graph facts and composition evidence, which newly added edge triples are individually necessary for \texttt{E68} to reach \texttt{E52}?
& \shortstack[l]{
    [[E23, R3, E42], [E42, R4, E61]]
} \\

\bottomrule
\end{tabular}
\end{table*}

This appendix describes the graph sources, view generation, question and answer construction, rendering, and validation procedures used in \textsc{MVGraph}.
Table~\ref{tab:advanced_task_examples}--\ref{tab:pure_reasoning_task_examples} illustrates the examples of each task.
We also detail how the stage-oriented diagnostic tasks are constructed from the same visual-graph setting to probe perception, alignment, composition, and reasoning.

\subsection{Construction Pipeline}

An instance is built by generating or sampling a task-conditioned graph structure, constructing interrelated views, instantiating a question and its programmatically computed answer, applying task-specific checks, and rendering the accepted views as images.
The views form an indexed sequence $\mathcal{I}=(I_1,\ldots,I_m)$, where $2\leq m\leq 8$.
The eight end-to-end graph-theoretic tasks and their structure-prepared reasoning conditions use two to four views; the remaining diagnostic tasks can use two to eight views.
In particular, MFC uses three or four temporal snapshots.
The order of the views is retained for temporal questions.

Each image $I_i$ is rendered from a graph $G_i=(V_i,\mathcal{E}_i)$, with node set $V_i$ and edge triples $\mathcal{E}_i\subseteq V_i\times\mathcal{R}\times V_i$.
The same node identifier denotes the same entity across views, but node positions are determined independently for each image.
Both node and edge sets may differ between views; shared coordinates are therefore neither assumed nor needed for cross-view identity matching.
For perception, alignment, and composition, the model receives the graph images.
In the perception protocol, it receives multiple images together and produces a separate textual node-and-edge representation for each view.
For reasoning diagnostics, the model instead receives task-prepared textual graph structure or composition evidence, as described below.

\subsection{Graph Sources and View Generation}

\paragraph{Task-conditioned synthetic graphs.}
Synthetic generators construct candidate graphs around the structural properties needed by each task rather than sampling arbitrary graphs and searching retrospectively for suitable questions.
For partial views, the generators control shared and view-specific nodes and edges, allocate complementary path segments, and add distractor structures.
For temporal views, they apply controlled node or edge changes to successive snapshots and select transitions or structures that satisfy the target question.
These procedures retain the original local-view construction rules while organizing the resulting diagnostic tasks according to their primary target capability.

\emph{Alignment} tasks control cross-view overlap or temporal change in nodes and complete edge triples.
The partial-view tasks are shared nodes (A-SN), view-specific nodes (A-VSN), shared edges (A-SE), and view-specific edges (A-VSE).
The temporal tasks are added nodes (A-AN), removed nodes (A-RN), added edges (A-AE), removed edges (A-RE), persistent nodes (A-PN), and persistent edges (A-PE).
Relation identifiers may occur on multiple edges, so edge identity is determined by the complete $(u,r,v)$ triple, not by $r$ alone.
Matching the same node across views relies on its shared identifier; the tasks do not introduce aliases or an additional entity-linking problem.

\emph{Composition} tasks require organizing evidence distributed across views.
For partial views, union node count (C-UN) and union edge count (C-UE) operate on the deduplicated union; path assembly (C-PA) requests a source--target path as an ordered sequence of edge triples, or \texttt{NONE} when no such path exists; and component decomposition (C-CD) requests every component of the union graph with its nodes and edges, including isolated nodes.
Minimal view set (C-MVS) identifies a smallest subset of views that preserves source--target reachability, whereas essential views (C-EV) identifies views whose individual removal destroys it.
Their generators distribute path evidence across views and introduce other structures so that the designated view contributions can be tested.
For temporal views, persistent path (C-PP) searches the intersection of the node and edge sets of all snapshots for a source--target path, returning \texttt{NONE} if absent.
Temporal path composition (C-TPC) constructs a transition at which initially separate source and target components first become connected; its answer contains the two complete pre-transition component subgraphs and a post-transition connecting path.

The structure-prepared \emph{reasoning} diagnostics reuse the eight graph-theoretic problems described below and additionally include bridge-node criticality (R-BN) and causal edge identification (R-CE).
R-BN uses an already composed graph and its cross-view shared-node list to test which shared nodes disconnect a designated source and target when removed.
R-CE uses ordered temporal structures and prepared composition evidence to identify newly added edges necessary for the queried reachability transition.
The diagnostic preparation changes the information supplied to the model, not the underlying graph-theoretic question.

\paragraph{SmallPedia topologies.}
To complement controlled synthetic structures with naturally occurring connectivity patterns, we also draw local topologies from the static and temporal components of \textsc{tkgl-smallpedia} in TGB~2.0~\citep{gastinger2024tgb} for applicable diagnostic instances.
SmallPedia supplies topology only; its original images, entity names, and semantic question--answer pairs are not reused.
We remove self-loops, treat sampled structures as undirected, and deterministically retain one relation when multiple facts connect the same unordered node pair.
Entities and relations are anonymized as \texttt{E*} and \texttt{R*}, and the resulting graphs are rendered in the same style as synthetic instances.
The sampled edges and anonymization mappings are retained in instance metadata for traceability.
For partial views, sampled local static structures are partitioned into overlapping subgraphs while preserving their sampled connectivity.
For temporal views, local structures are constructed from timestamped facts in their original order; a transition question identifies the queried pair of consecutive snapshots.
Questions and answers are generated from these transformed anonymous graphs, not from the original entity semantics.

\paragraph{Graph-theoretic challenge generators.}
The eight end-to-end tasks are generated using randomized procedures specialized to four graph-theoretic problem families.
For the partial-view tasks, a complete problem graph is generated and validated before its edges are distributed across two to four views.
Edge allocation is balanced, and each view's node set is determined by its incident edges and additional nodes sampled from the complete graph.
For the temporal tasks, successive graphs contain persistent and time-specific nodes, edges, or capacities.
Candidates are filtered by exact algorithms and problem-specific non-degeneracy constraints.

\emph{Topological sorting.}
For TS, we generate a directed acyclic graph with 12--14 nodes and distribute its edges over two to four weakly connected local views.
Each view contributes ordering constraints not fully implied by the other views, and the complete graph admits multiple valid topological orders while retaining a sufficiently long dependency chain.
For STS, each snapshot contains seven stable nodes and one time-specific node.
The answer gives a valid order for each snapshot while minimizing the total Kendall inversion distance between the relative orders of nodes shared by consecutive snapshots.

\emph{Maximum flow.}
For MF, a directed capacity network with designated source and sink is distributed across two to four partial views.
Candidates are retained only when removing any view decreases the maximum flow of the composed network.
For MFC, three or four capacity-network snapshots contain nine stable core nodes and one or two time-specific nodes.
Changes to edges and capacities yield different source--sink maximum-flow values, and the answer identifies the consecutive pair with the largest flow increase.

\emph{Bipartite matching.}
For BM, a bipartite graph with 8--10 applicant nodes and 8--10 job nodes is distributed across two to four partial views.
The generator introduces a controlled Hall deficiency and preserves replaceable matching edges, so the maximum-cardinality solution need not match every applicant or be unique.
For SBM, each snapshot contains seven stable applicants, seven stable jobs, one time-specific applicant, and one time-specific job.
Candidate edges vary across time; the answer attains maximum matching cardinality in every snapshot while minimizing the number of job changes by stable applicants between consecutive snapshots.

\emph{Hamiltonian paths.}
For HP, connected undirected graphs of 14--17 nodes are filtered using exact memoized Hamiltonian-path search before their edges are distributed across two to four views.
Candidates without a Hamiltonian path or with trivial degree patterns are rejected.
For THP, two to four temporal views have different node and edge sets, with 7--11 nodes in each view.
An accepted sequence admits a Hamiltonian path whose edge-image indices are nondecreasing and collectively use every time step.
The answer contains both the node sequence and the image index assigned to each path edge.

\subsection{Graph Representation and Rendering}

The partial-view and temporal diagnostic graphs derived from the original foundational generators, including their SmallPedia-derived counterparts, are undirected.
Among the graph-theoretic tasks, topological sorting and maximum flow use directed graphs, whereas bipartite matching and Hamiltonian paths use undirected graphs.
Nodes have semantically neutral \texttt{E*} identifiers and reusable relation types have \texttt{R*} identifiers; maximum-flow edges additionally display positive numerical capacities.
These labels prevent domain-specific lexical semantics from providing the answer.

Each rendered view contains 7--20 nodes, a bound applied to individual images rather than to the union of all views.
Layouts are computed independently from local graph structures, so recurring nodes can appear at different positions.
Node shapes, fonts, edge styles, relation labels, arrowheads where applicable, and image titles follow a unified rendering style.
Graph-theoretic layouts are additionally checked for node spacing and, for maximum-flow networks, overlap involving capacity labels.
Layouts that fail the applicable visual checks are regenerated.

\subsection{Question and Answer Generation}

Task-specific templates instantiate questions after the graph views have been finalized.
The output format depends on the task: answers can be counts, node or edge-triple lists, view indices, component subgraphs, paths, topological orders, matchings, or task-specific temporal outputs.
The perception protocol instead requests a separate textual representation of the nodes and edge triples in every input image.
All reference answers are computed from the underlying graph data rather than from model-generated annotations.

For alignment, answers are obtained through node-set or edge-triple intersection and difference across partial views or designated temporal snapshots.
For composition, the construction computes deduplicated unions, components, path witnesses, persistent structures, or sufficient and indispensable view subsets as specified by the task.
For end-to-end graph-theoretic tasks, exact graph algorithms compute reference orders, flow values, matchings, and Hamiltonian paths, including temporal stability or edge-time constraints where applicable.
R-BN and R-CE use the prepared structural evidence to formulate node-removal and newly added-edge necessity queries, respectively.

The reasoning-diagnostic input is prepared from the same underlying structures.
For the four partial-view graph-theoretic tasks, the graph is already composed into a textual representation; for temporal tasks, the snapshot sequence and its order are preserved.
R-BN additionally supplies the composed graph and the list of shared candidate nodes, whereas R-CE supplies the temporal structures and composition evidence needed to locate the relevant transition.
Thus these conditions target graph-theoretic and counterfactual inference after the task-relevant structure has been made explicit.

Several graph-theoretic questions admit multiple correct outputs, so the stored answer is a reference rather than the sole acceptable response.
The topological-order verifier checks node coverage, uniqueness, and precedence constraints.
The matching verifier checks partition membership, edge validity, one-to-one assignments, and maximum cardinality.
The Hamiltonian-path verifier checks node coverage, uniqueness, adjacency, and, for THP, nondecreasing and complete coverage of the required edge-image indices.
For STS and SBM, the verifier additionally checks the respective minimum inversion or reassignment cost across snapshots.

\subsection{Evidence Necessity and Validation}

For tasks intended to test cross-view dependence, a view is treated as required when removing it changes the programmatically established answer or leaves insufficient information to establish that answer under the task semantics.
Where available, required-view indices are retained in the metadata and are not supplied to the evaluated model.
When a question remains well-defined after removing a view, leave-one-out answer recomputation compares the reduced-view and full-view answers.
For tasks such as intersections, persistent properties, and negative reachability, unchanged recomputed values alone do not establish that a view was redundant, because the omitted view may be needed to rule out alternatives.
The perception protocol is different: it tests reconstruction of each supplied view and does not impose a cross-view evidence-necessity criterion on its output.

Task-specific construction checks also enforce the intended structural dependencies.
For example, TS requires independent ordering constraints across views, MF requires every view to contribute to the maximum-flow value, and temporal tasks retain the snapshots needed to establish their queried transition or persistent property.
Diagnostic tasks use their corresponding overlap, path-distribution, temporal-change, and evidence-selection constraints rather than a single universal leave-one-out test.

Across tasks, structural checks require edge endpoints to belong to their view's node set, exclude disallowed self-loops and duplicate edges, and enforce the prescribed per-view node bounds.
Task-specific checks verify properties such as acyclicity, bipartite structure, transition direction, path existence, and solution feasibility where relevant.
Stored reference answers are checked against the finalized graph structures, and tasks with multiple valid solutions use the semantic verifiers described above.
Non-degeneracy rules are applied by task, rather than rejecting empty outputs globally; in particular, C-PA and C-PP can legitimately return \texttt{NONE}.
Finally, views that fail applicable readability checks, including node-spacing and label-placement constraints, are regenerated.

\begin{table*}[t]
\centering
\caption{Per-task statistics for the end-to-end graph-theoretic challenges.
Average node and edge counts are computed after deduplication across all views
of an instance.}
\label{tab:app_graph_theoretic_stats}
\footnotesize
\setlength{\tabcolsep}{4pt}
\renewcommand{\arraystretch}{1.08}
\begin{tabular}{llrrrrr}
\toprule
\textbf{ID} & \textbf{Task} & \textbf{Instances} &
\textbf{Views} & \textbf{Avg. views} &
\textbf{Avg. nodes} & \textbf{Avg. edges} \\
\midrule
\rowcolor{gray!10}\multicolumn{7}{l}{\emph{Partial graph views}} \\
\rowcolor{white}
TS & Topological Sorting & 1,000 & 2--4 & 2.98 & 13.01 & 36.23 \\
MF & Maximum Flow & 1,000 & 2--4 & 2.98 & 12.00 & 25.54 \\
BM & Bipartite Matching & 1,000 & 2--4 & 3.02 & 18.00 & 34.28 \\
HP & Hamiltonian Path & 1,000 & 2--4 & 3.03 & 15.51 & 40.91 \\
\addlinespace[2pt]
\rowcolor{gray!10}\multicolumn{7}{l}{\emph{Temporal graph views}} \\
\rowcolor{white}
STS & Stable Topological Sorting & 1,000 & 2--4 & 3.03 & 10.03 & 32.01 \\
MFC & Maximum-flow Change & 1,000 & 3--4 & 3.49 & 14.19 & 24.47 \\
SBM & Stable Bipartite Matching & 1,000 & 2--4 & 2.98 & 19.95 & 50.79 \\
THP & Time-respecting Hamiltonian Path & 1,000 & 2--4 & 2.97 & 15.49 & 28.02 \\
\midrule
\textbf{Total} & & \textbf{8,000} & \textbf{2--4} &
\textbf{3.06} & \textbf{14.77} & \textbf{34.03} \\
\bottomrule
\end{tabular}
\end{table*}

\begin{table*}[t]
\centering
\caption{Per-task statistics for the alignment diagnostics.}
\label{tab:app_alignment_stats}
\footnotesize
\setlength{\tabcolsep}{4pt}
\renewcommand{\arraystretch}{1.08}
\begin{tabular}{llrrrrr}
\toprule
\textbf{ID} & \textbf{Task} & \textbf{Instances} &
\textbf{Views} & \textbf{Avg. views} &
\textbf{Avg. nodes} & \textbf{Avg. edges} \\
\midrule
\rowcolor{gray!10}\multicolumn{7}{l}{\emph{Partial graph views}} \\
\rowcolor{white}
A-SN  & Shared Nodes & 1,000 & 2--8 & 4.78 & 33.10 & 47.51 \\
A-VSN & View-Specific Nodes & 1,000 & 2--8 & 4.76 & 32.91 & 40.19 \\
A-SE  & Shared Edges & 1,000 & 2--8 & 4.72 & 33.90 & 44.91 \\
A-VSE & View-Specific Edges & 1,000 & 2--8 & 4.72 & 32.71 & 41.10 \\
\addlinespace[2pt]
\rowcolor{gray!10}\multicolumn{7}{l}{\emph{Temporal graph views}} \\
\rowcolor{white}
A-AN & Added Nodes & 1,000 & 3--8 & 4.98 & 10.00 & 13.14 \\
A-RN & Removed Nodes & 1,000 & 3--8 & 4.99 & 9.99 & 13.28 \\
A-AE & Added Edges & 1,000 & 3--8 & 4.95 & 13.43 & 10.58 \\
A-RE & Removed Edges & 1,000 & 3--8 & 5.00 & 13.45 & 10.57 \\
A-PN & Persistent Nodes & 1,000 & 3--8 & 4.99 & 34.63 & 42.41 \\
A-PE & Persistent Edges & 1,000 & 3--8 & 5.00 & 26.56 & 21.21 \\
\midrule
\textbf{Total} & & \textbf{10,000} & \textbf{2--8} &
\textbf{4.89} & \textbf{24.07} & \textbf{28.49} \\
\bottomrule
\end{tabular}
\end{table*}

\begin{table*}[t]
\centering
\caption{Per-task statistics for the composition diagnostics.}
\label{tab:app_composition_stats}
\footnotesize
\setlength{\tabcolsep}{4pt}
\renewcommand{\arraystretch}{1.08}
\begin{tabular}{llrrrrr}
\toprule
\textbf{ID} & \textbf{Task} & \textbf{Instances} &
\textbf{Views} & \textbf{Avg. views} &
\textbf{Avg. nodes} & \textbf{Avg. edges} \\
\midrule
\rowcolor{gray!10}\multicolumn{7}{l}{\emph{Partial graph views}} \\
\rowcolor{white}
C-UN  & Union Node Count & 1,000 & 2--8 & 4.71 & 32.51 & 39.72 \\
C-UE  & Union Edge Count & 1,000 & 2--8 & 4.70 & 32.70 & 39.67 \\
C-PA  & Path Assembly & 1,000 & 2--8 & 4.75 & 29.48 & 24.91 \\
C-CD  & Component Decomposition & 1,000 & 2--8 & 4.75 & 12.73 & 10.22 \\
C-MVS & Minimal View Set & 1,000 & 2--8 & 4.79 & 31.41 & 26.45 \\
C-EV  & Essential Views & 1,000 & 2--8 & 4.71 & 30.90 & 26.02 \\
\addlinespace[2pt]
\rowcolor{gray!10}\multicolumn{7}{l}{\emph{Temporal graph views}} \\
\rowcolor{white}
C-PP  & Persistent Path & 1,000 & 3--8 & 5.00 & 18.91 & 17.91 \\
C-TPC & Temporal Path Composition & 1,000 & 3--8 & 5.04 & 14.02 & 12.48 \\
\midrule
\textbf{Total} & & \textbf{8,000} & \textbf{2--8} &
\textbf{4.81} & \textbf{25.33} & \textbf{24.67} \\
\bottomrule
\end{tabular}
\end{table*}

\begin{table*}[t]
\centering
\caption{Per-task statistics for the structure-prepared reasoning diagnostics.
The eight graph-theoretic tasks reuse the underlying instances in
Table~\ref{tab:app_graph_theoretic_stats}, but supply structural information
rather than requiring visual graph reconstruction.}
\label{tab:app_reasoning_stats}
\footnotesize
\setlength{\tabcolsep}{4pt}
\renewcommand{\arraystretch}{1.08}
\begin{tabular}{llrrrrr}
\toprule
\textbf{ID} & \textbf{Task} & \textbf{Instances} &
\textbf{Views} & \textbf{Avg. views} &
\textbf{Avg. nodes} & \textbf{Avg. edges} \\
\midrule
\rowcolor{gray!10}\multicolumn{7}{l}{\emph{Partial graph views}} \\
\rowcolor{white}
R-TS & Topological Sorting & 1,000 & 2--4 & 2.98 & 13.01 & 36.23 \\
R-MF & Maximum Flow & 1,000 & 2--4 & 2.98 & 12.00 & 25.54 \\
R-BM & Bipartite Matching & 1,000 & 2--4 & 3.02 & 18.00 & 34.28 \\
R-HP & Hamiltonian Path & 1,000 & 2--4 & 3.03 & 15.51 & 40.91 \\
R-BN & Bridge-Node Criticality & 1,000 & 2--8 & 4.78 & 29.69 & 25.08 \\

\addlinespace[2pt]
\rowcolor{gray!10}\multicolumn{7}{l}{\emph{Temporal graph views}} \\
\rowcolor{white}
R-STS & Stable Topological Sorting & 1,000 & 2--4 & 3.03 & 10.03 & 32.01 \\
R-MFC & Maximum-flow Change & 1,000 & 3--4 & 3.49 & 14.19 & 24.47 \\
R-SBM & Stable Bipartite Matching & 1,000 & 2--4 & 2.98 & 19.95 & 50.79 \\
R-THP & Time-respecting Hamiltonian Path & 1,000 & 2--4 & 2.97 & 15.49 & 28.02 \\
R-CE & Causal Edge Identification & 1,000 & 3--8 & 5.05 & 14.02 & 12.51 \\
\midrule
\textbf{Total} & & \textbf{10,000} & \textbf{2--8} &
\textbf{3.43} & \textbf{16.19} & \textbf{30.98} \\
\bottomrule
\end{tabular}
\end{table*}

\subsection{Dataset Organization and Scale}

Each instance record stores the rendered graph views, its question and gold
answer, per-view graph structures in JSON, and instance-level metadata.
The metadata identifies the task, image paths, and answer format, and records
construction and validation information where applicable.
For SmallPedia-derived instances, the source edges and anonymization mappings
are also retained.
Models do not receive the underlying graph JSON or construction metadata in
image-based evaluation.
In the structure-prepared reasoning condition, textualized structural evidence
is instead supplied to probe downstream reasoning.

\textsc{MVGraph} comprises 37 evaluation tasks and 72,000 task-instance
records across five data collections.
The eight end-to-end graph-theoretic tasks contribute 8,000 records.
The diagnostic collections comprise 36,000 perception, 10,000 alignment,
8,000 composition, and 10,000 reasoning records.
Perception uses a single evaluation objective, reconstructing the nodes and
edge triples of each supplied graph view, across 36,000 multi-view instances
with 2--8 views and an average of 4.46 views.
These instances contain, on average, 21.47 distinct nodes and 26.93 distinct
edge triples across their views.
Tables~\ref{tab:app_graph_theoretic_stats}--\ref{tab:app_reasoning_stats}
provide per-task statistics for the remaining four groups.

\section{Inference Prompts and Instructions}
\label{app:common-prompts}

This appendix summarizes the prompts and instructions provided to models
during inference. Each input combined shared stage-level instructions
with sample-specific graph context, a task question, and an output
requirement. Bracketed placeholders indicate content instantiated for
each sample; they were not shown literally to the model.

\subsection{End-to-End Graph-Theoretic Reasoning}
\label{app:rh-prompts}

The end-to-end tasks received the
ordered graph images directly. Each sample used the following structure:

\begin{promptbox}{End-to-end prompt structure}
\textit{[Scenario statement]}

\medskip
\textit{[Task-specific question]}

\medskip
Return only a valid JSON object of the form
\textit{[task-specific answer schema]}.
Do not include explanation or extra fields.
\end{promptbox}

The scenario statement specified whether the images were partial views
or temporal snapshots, whether edges were directed, and whether edge
labels represented relation types or numeric capacities. Identical
\texttt{E*} labels denoted the same node across images. For
non-capacity tasks, edges were triples
[head node, relation type, tail node], where \texttt{R*} was a
reusable relation type. For capacity tasks, the displayed number on
each directed edge was its capacity; relation labels were not shown.

\begin{promptbox}{End-to-end task questions}
\begin{description}
\item[\textbf{TS (RH1).}]
Merge all views into one directed acyclic graph. Give a valid
topological ordering of every node in the merged graph.
The answer is a JSON list of node labels.

\item[\textbf{STS (RH2).}]
Give one valid topological ordering for each time step, in image
order. Among all valid sequences, minimize the total number of
pairwise order inversions for nodes shared by consecutive time steps.
The answer is a JSON list of node-label lists.

\item[\textbf{MF (RH3).}]
Merge all views into one capacity network. The source is
\textit{[source]} and the sink is \textit{[sink]}. What is the
maximum-flow value from source to sink?
The answer is a JSON integer.

\item[\textbf{MFC (RH4).}]
The source is \textit{[source]} and the sink is \textit{[sink]}.
Between which consecutive graph images does the maximum-flow value
increase the most? Return the two graph image numbers in chronological
order. The answer is a JSON list of two image numbers.

\item[\textbf{BM (RH5).}]
The applicant partition is \textit{[applicants]}, and the job partition
is \textit{[jobs]}. Find a maximum-cardinality assignment in the
merged graph. Return applicant-job pairs.
The answer is a JSON list of pairs.

\item[\textbf{SBM (RH6).}]
The applicant partitions in image order are
\textit{[applicants by time]}, and the job partitions are
\textit{[jobs by time]}. Return a maximum-cardinality matching for
every time step. Among all such sequences, minimize how many stable
applicants change their assigned job between consecutive images.
The answer is a JSON list of matchings.

\item[\textbf{HP (RH7).}]
Merge all views into one undirected graph. Give a Hamiltonian path
that visits every node in the merged graph exactly once.
The answer is a JSON list of node labels.

\item[\textbf{THP (RH8).}]
Find a time-respecting Hamiltonian path that visits every node
exactly once. For each path edge, give the graph image in which
that edge is used. Image numbers must be nondecreasing, and every
time step must be used. The answer is a JSON object containing
\texttt{path} and \texttt{images} lists.
\end{description}
\end{promptbox}

\subsection{Perception}

Models received the graph images in order. The common image-parsing
instruction was:

\begin{promptbox}{Perception prompt}
You are a graph image parsing assistant.

\medskip
\textit{[Graph scenario description]}

\medskip
Inspect all graph images and reconstruct every graph shown. Do not infer
relation meanings or facts that are not visible in the images.

\medskip
Return exactly one JSON object containing a \texttt{graphs} list.
For each graph, preserve its \texttt{graph\_id} and output its
\texttt{triples}. Extract every node, including isolated nodes;
represent an isolated node as
\texttt{["E80", "", ""]}. Preserve node and edge labels exactly.
Do not include explanations, Markdown, or additional keys.
\end{promptbox}

The undirected variant emitted each edge once. The directed variant
preserved arrow direction without adding reverse edges. For capacity
networks, the image-parsing output used the visible numeric capacity
rather than inventing an unshown relation label.

\subsection{Alignment and Composition}

These image-based tasks shared a scenario preamble but had no separate
stage-specific system prompt. For partial views, the preamble was:

\begin{promptbox}{Partial-view scenario preamble}
The images show partial views of related undirected graphs. Same
\texttt{E*} labels indicate the same node across views, and edges are
triples [head node, relation type, tail node] where \texttt{R*} is the
reusable relation type.
\end{promptbox}

For temporal views, it was:

\begin{promptbox}{Temporal-view scenario preamble}
The images show an undirected graph at different time steps. Same
\texttt{E*} labels indicate the same node across time, and edges are
triples [head node, relation type, tail node] where \texttt{R*} is the
reusable relation type.
\end{promptbox}

Each preamble was followed by the task-specific question and its
JSON output instruction. Alignment questions requested shared,
view-specific, added, removed, or persistent nodes and edge triples.
Composition questions requested union statistics, assembled paths,
component subgraphs, necessary views, or temporal structural evidence.
The required JSON schema depended on the task; it was not uniformly
a node list.

\subsection{Reasoning}

Models received prepared graph facts rather than images. The common
instruction began:

\begin{promptbox}{Reasoning prompt}
You are a graph reasoning assistant.

\medskip
Answer the question using only the graph facts and question provided
below. Do not use images, outside knowledge, or unstated assumptions.

\medskip
Graph text construction:
\textit{[Scenario, CSV columns, edge semantics, and isolated-node
encoding.]}
Treat node labels and relation labels as exact strings. Relation labels
such as \texttt{R1}, \texttt{R2}, \texttt{R3}, and \texttt{R4} have no
implicit meaning.

\medskip
Graph facts:
\textit{[Graph CSV]}

\medskip
Question:
\textit{[Task-specific question and JSON output instruction]}
\end{promptbox}

For partial-view reasoning tasks, the supplied facts contained the
merged graph. Temporal tasks retained separate snapshots in time order.
The CSV description varied with edge direction and capacity; the
causal-edge task additionally supplied Composition evidence.

\section{End-to-End Graph-Theoretic Performance}
\subsection{Main Results}
\label{app:main_results}

\begin{table}[t]
\centering
\small
\setlength{\tabcolsep}{3pt}
\caption{Performance of LVLMs on the 8 end-to-end graph-theoretic reasoning tasks of \textsc{MVGraph}, where $\dagger$ denotes evaluation via API called. The best results are \textbf{bolded}.}
\label{tab:graph_theoretic_all}
\begin{tabular}{lrrrrrrrrr}
\toprule
\textbf{Model} & \textbf{TS} & \textbf{STS} & \textbf{MF} & \textbf{MFC} & \textbf{BM} & \textbf{SBM} & \textbf{HP} & \textbf{THP}  & \textbf{Avg.}\\
\midrule
\rowcolor{gray!10}
\multicolumn{10}{@{}l}{\textit{No-Thinking}} \\
\rowcolor{white}
Gemma-4-12B-it & \textbf{0.0} & \textbf{0.0} & 0.0 & 40.9 & 0.0 & \textbf{0.1} & \textbf{4.0} & 0.0 & 5.6 \\
Qwen3.5-9B & \textbf{0.0} & \textbf{0.0} & 0.0 & 41.5 & 0.0 & 0.0 & 0.2 & 0.0 & 5.2 \\
Ovis2.5-9B & \textbf{0.0} & \textbf{0.0} & 0.0 & 41.7 & 0.0 & 0.0 & 0.3 & 0.0 & 5.2 \\
Qwen3-VL-8B & \textbf{0.0} & \textbf{0.0} & 0.0 & 42.3 & 0.0 & 0.0 & 1.0 & 0.0 & 5.4 \\
GLM-4.6V-Flash & \textbf{0.0} & \textbf{0.0} & 0.0 & 41.7 & 0.0 & 0.0 & 0.0 & 0.0 & 5.2 \\
Molmo2-8B & \textbf{0.0} & \textbf{0.0} & \textbf{2.8} & 42.5 & 0.0 & 0.0 & 0.0 & 0.0 & \textbf{5.7} \\
MiniCPM-V-4.5 & \textbf{0.0} & \textbf{0.0} & 1.0 & 31.5 & 0.0 & 0.0 & 0.2 & 0.0 & 4.1 \\
Pixtral-12B & \textbf{0.0} & \textbf{0.0} & 0.0 & 15.4 & 0.0 & 0.0 & 0.0 & 0.0 & 1.9 \\
InternVL3.5-8B & \textbf{0.0} & \textbf{0.0} & 0.0 & \textbf{42.7} & \textbf{0.1} & 0.0 & 0.0 & 0.0 & 5.4 \\
LLaVA-OneVision-2-8B & \textbf{0.0} & \textbf{0.0} & 0.0 & 42.0 & 0.0 & 0.0 & 0.2 & \textbf{0.1} & 5.3 \\
Phi-4-reasoning-vision-15B & \textbf{0.0} & \textbf{0.0} & 0.1 & 42.6 & 0.0 & 0.0 & 0.1 & 0.0 & 5.4 \\
Llama-3.1-Nemotron-Nano-VL-8B & \textbf{0.0} & \textbf{0.0} & 0.0 & 40.6 & 0.0 & 0.0 & 0.0 & 0.0 & 5.1 \\
\cmidrule(lr){1-10}
DeepSeek-Flash$^\dagger$ & \textbf{0.0} & \textbf{0.0} & \textbf{0.0} & \textbf{45.0} & \textbf{2.0} & \textbf{0.0} & \textbf{2.0} & \textbf{2.0} & \textbf{6.4} \\
Qwen3.8-Flash$^\dagger$ & \textbf{0.0} & \textbf{0.0} & \textbf{0.0} & 17.0 & \textbf{2.0} & \textbf{0.0} & \textbf{2.0} & 0.0 & 2.6 \\
\midrule
\rowcolor{gray!10}
\multicolumn{10}{@{}l}{\textit{Thinking}} \\
\rowcolor{white}
Qwen3.5-9B & \textbf{0.1} & 0.0 & 0.3 & 30.2 & 36.9 & 0.2 & 40.0 & 0.4 & 13.5 \\
Gemma-4-12B-it & 0.0 & \textbf{0.1} & \textbf{5.5} & 45.3 & \textbf{60.6} & \textbf{0.4} & \textbf{43.1} & \textbf{8.7} & \textbf{20.5} \\
GLM-4.6V-Flash & 0.0 & 0.0 & 0.7 & \textbf{46.5} & 7.2 & 0.0 & 4.7 & 0.0 & 7.4 \\
Ovis2.5-9B & 0.0 & 0.0 & 0.2 & 41.3 & 18.5 & 0.0 & 2.3 & 0.2 & 7.8 \\
\cmidrule(lr){1-10}
Qwen3.8-Flash$^\dagger$ & \textbf{35.0} & 8.0 & \textbf{54.0} & \textbf{59.0} & \textbf{97.0} & 8.0 & 77.0 & 65.0 & 50.4 \\
GLM-5.3-Flash$^\dagger$ & 33.0 & \textbf{12.0} & 41.0 & \textbf{59.0} & 95.0 & \textbf{16.0} & \textbf{84.0} & \textbf{79.0} & \textbf{52.4} \\
DeepSeek-Flash$^\dagger$ & 0.0 & 0.0 & 15.0 & 5.0 & 62.0 & 0.0 & 50.0 & 7.0 & 17.4 \\
\bottomrule
\end{tabular}
\end{table}

Table~\ref{tab:graph_theoretic_all} reports the complete results across
inference settings.
Without thinking, all 14 evaluated models achieve low average accuracy,
ranging from 1.9\% to 6.4\%.
Topological sorting (TS) and stable topological sorting (STS) receive no
correct answers, while performance on most other tasks is also near zero.
Maximum-flow change (MFC) is the main exception, with substantially higher
scores for many models.
These scores warrant caution: MFC asks models to select a pair of consecutive
snapshots from a sequence of three or four, rather than to output or justify
the maximum-flow values themselves.
Its smaller answer space means that correct selections do not necessarily
demonstrate reliable flow computation, and MFC contributes disproportionately
to the no-thinking averages.

Thinking improves the average accuracy of every model evaluated in both
settings, but the magnitude of the improvement varies considerably.
Qwen3.8-Flash rises from 2.6\% to 50.4\%, and Gemma-4-12B-it rises from
5.6\% to 20.5\%, whereas the gains for GLM-4.6V-Flash and Ovis2.5-9B are
much smaller.
Among the thinking models, GLM-5.3-Flash and Qwen3.8-Flash perform best
overall, followed at a substantial distance by the strongest locally
evaluated model, Gemma-4-12B-it.
DeepSeek-Flash performs below Gemma-4-12B-it despite being API-accessed,
showing that the performance differences cannot be characterized simply
by deployment category.
These comparisons are descriptive rather than size-controlled, and
GLM-5.3-Flash has no corresponding no-thinking result in this evaluation.

Performance also depends strongly on the graph-theoretic problem.
The two leading models achieve 95.0--97.0\% on bipartite matching (BM),
but only 8.0--16.0\% on stable bipartite matching (SBM).
Their TS scores likewise exceed those on STS.
The temporal gap is not uniform, however: both models retain comparatively
strong performance on time-respecting Hamiltonian path (THP).
Moreover, each static--temporal task pair differs in construction, output
requirements, or both, so these score differences should not be interpreted
as an isolated effect of temporal reasoning.
Taken together, the full results show that additional test-time reasoning
can yield large gains for some models and tasks, while leaving major
graph-theoretic challenges unresolved.
They motivate the stage-oriented diagnostics used to examine which
capabilities limit end-to-end performance.

\end{document}